\documentclass[10pt, a4paper]{dukenlp}

\usepackage{times}
\usepackage{latexsym}
\usepackage{natbib}
\usepackage[T1]{fontenc}

\usepackage[utf8]{inputenc}

\usepackage{microtype}

\usepackage{inconsolata}

\usepackage{graphicx}

\usepackage{tabularx}
\usepackage{xcolor, colortbl}
\usepackage{booktabs}
\usepackage[most]{tcolorbox}
\usepackage{cleveref}
\usepackage{wrapfig}
\definecolor{darkblue}{rgb}{0, 0, 0.5}
\definecolor{darkblue}{RGB}{0,51,102}

\tcbset{
  promptbox/.style={
    enhanced,
    colback=gray!3,
    colframe=darkblue,
    boxrule=0.5pt,
    left=6pt, right=6pt, top=6pt, bottom=6pt,
    fontupper=\ttfamily\footnotesize,
    before upper=\setlength{\parskip}{3pt}\setlength{\parindent}{0pt},
    sharp corners,
    breakable,
    before skip=6pt, after skip=6pt,
  }
}

\newtcolorbox{promptbox}[1][]{%
  promptbox,
  title={#1},
  coltitle=white,
  fonttitle=\sffamily\small\bfseries,
  attach boxed title to top left={yshift=-2mm,xshift=4mm},
  boxed title style={colback=darkblue,boxrule=0.5pt,sharp corners},
}
\title{Rethinking Inference-Time Scaling: Efficiency Limits and Linguistic Signals}

\author{%
  \textbf{Junlin Wang}\textsuperscript{1}\thanks{Work done while interning at Together AI},
  \textbf{Shang Zhu}\textsuperscript{2},
  \textbf{Jon Saad-Falcon}\textsuperscript{2,4},
  \textbf{Ben Athiwaratkun}\textsuperscript{2},
  \textbf{Qingyang Wu}\textsuperscript{2} \\
  \textbf{Jue Wang\textsuperscript{2},}
  \textbf{Shuaiwen Leon Song\textsuperscript{2},}
  \textbf{Ce Zhang\textsuperscript{2,3},}
  \textbf{Bhuwan Dhingra\textsuperscript{1},}
  \textbf{James Zou\textsuperscript{2,4}}
  \\
  \textsuperscript{1}Duke University,
  \textsuperscript{2}Together AI,
  \textsuperscript{3}University of Chicago,
  \textsuperscript{4}Stanford University \\
  {\tt\small junlin.wang2@duke.edu,\, \{shang, ben, qingyang, jue, leon\}@together.ai} \\
  {\tt\small cez@uchicago.edu, \{jonsaadfalcon, jamesz\}@stanford.edu, bdhingra@cs.duke.edu}
}

\begin{document}
\maketitle
\begin{abstract}
There is intense interest in investigating how inference time compute (ITC) (e.g. repeated sampling, refinements, etc) can improve large language model (LLM) capabilities. While breakthroughs like DeepSeek-R1 highlight the power of reinforcement learning for reasoning, the interaction between ITC and reasoning-optimized weights remains poorly understood. This work conducts a comprehensive analysis of inference-time scaling methods for both reasoning and non-reasoning models on challenging reasoning tasks. While prior work suggests that scaling test-time compute can optimally substitute for model parameter scaling, we identify a fundamental limit to this compute-equivalence, the ``reasoning floor'', a performance plateau that non-reasoning models cannot escape, no matter how much inference compute is spent. 
We demonstrate that general-purpose models fail to match the accuracy of reasoning-optimized models even with an order of magnitude more inference compute, suggesting that internalizing reasoning protocols is a prerequisite for effective test-time scaling. Within reasoning models, we find that the complexity of the scaling method often yields diminishing returns; simple majority voting consistently outperforms sophisticated sequential revision and mixture-of-agents frameworks. Crucially, we identify a ``Linguistic Signal of Correctness''—correct responses are significantly more concise and exhibit a lower density of ``hedging'' and ``thinking'' markers. We demonstrate that these intrinsic linguistic features can serve as zero-compute proxies for response quality, providing a pathway to more efficient, self-diagnostic reasoning agents.
\end{abstract}

\section{Introduction}
Language models have witnessed remarkable advancements in recent years, demonstrating increasingly sophisticated capabilities across various tasks \citep{openai2023gpt4,dubey2024llama3herdmodels,qwen}. Despite these improvements, complex reasoning remains challenging, often requiring additional computational resources and specialized techniques to achieve satisfactory performance \citep{wang-etal-2024-reasoning-token,tot}. This challenge has motivated the development of inference-time compute (ITC) scaling methods, which allocate additional computational resources during inference to enhance model outputs.

The landscape of language model reasoning has evolved along two primary dimensions. First, approaches like Chain-of-thought \citep{wei2023chainofthoughtpromptingelicitsreasoning}, self-consistency \citep{wang2022self0consistency}, tree-structured sampling \citep{snell2024scaling}, and mixture of agents \citep{wang2025mixtureofagents} have emerged as effective techniques for boosting reasoning capabilities during inference without requiring model parameter changes. Second, a new class of "reasoning models", explicitly post-trained to solve highly challenging problems, has been introduced, exemplified by models like o1 \citep{openai2024openai}, Deepseek-R1 \citep{deepseek-ai2025deepseek0r10}, and QwQ \citep{qwq-32b-preview}.

While both approaches show promise, they introduce significant computational overhead. Chain-of-thought \cite{wei2023chainofthoughtpromptingelicitsreasoning} prompting increases token generation, tree-structured sampling requires exploring multiple solution paths \citep{liu20251b}, and mixture of agents \citep{wang2025mixtureofagents} demands running several specialized agent configurations simultaneously. This computational burden raises critical questions about efficiency: how can we optimize the trade-off between computational resources and reasoning performance? Which inference-time scaling methods deliver the best results for different model architectures? How do reasoning models compare to conventional models under varying computational budgets?

These questions remain largely unsolved in the current literature. Prior work has primarily focused on fine-tuning strategies \citep{yu2025dapo0,xie2025logicrl} or simple inference evaluations \citep{brown2024large}, with limited attention to systematic evaluation of inference-time compute scaling across model types. Furthermore, a line of works suggest that inference-time compute can be more effective than model parameter scaling \citep{snell2024scaling}---these findings were largely established prior to the widespread availability of specialized reasoning architectures like DeepSeek-R1. Consequently, there is a lack of systematic evidence regarding whether this equivalence holds across different model classes. Furthermore, existing research often relies on resource-heavy process-reward models (PRMs), leaving the potential of reward-model-free (RM-free) scaling methods relatively misunderstood. 


Our work bridges this gap by conducting a comprehensive analysis of RM-free inference-time scaling across both reasoning-optimized and general-purpose architectures. We specifically focus on trained-verifier-free approaches to investigate the upper bounds of intrinsic model reasoning without external supervision. Our results challenge the compute-equivalence narrative, revealing a fundamental ``reasoning floor'': a performance plateau that non-reasoning models cannot escape, no matter how much inference compute is spent. On the other hand, for reasoning-optimized models, we find that increasing method complexity (e.g., sequential revision) often yields diminishing returns compared to simple majority voting. Our research provides the following key contributions:

\begin{itemize}
    \item  We compare inference-time scaling methods across reasoning and non-reasoning models, establishing efficiency–performance trade-off curves. We find that complex inference-time strategies yield diminishing returns, with simple majority voting consistently outperforming more elaborate methods.
    \item We provide empirical evidence that inference-time compute is not a universal substitute for specialized training. We show that non-reasoning models are fundamentally limited by a "reasoning floor," where even an order of magnitude more compute cannot match the performance of reasoning-optimized weights.
    \item We analyze the correlation between response linguistic features (response length, linguistic markers) and task performances, providing practical guidance for improving existing inference-time methods without increasing computational costs.

\end{itemize}


\section{Related Work}

\subsection{Inference-Time Scaling}

Recent work has demonstrated that scaling compute during inference offers a promising alternative to costly model pretraining. Language models \citep{touvron2023llamaopenefficientfoundation, jiang2023mistral, geminiteam2024geminifamilyhighlycapable} continue to improve with increased data and parameters, though at escalating development costs. 
Inference-time scaling approaches like \citet{brown2024largelanguagemonkeysscaling} show a log-linear relationship between problem-solving coverage and sample count across reasoning tasks, offering cost-effective alternatives to model size scaling.
Inference-time architectures combine techniques such as generation ensembling, sampling, ranking, and fusion to exceed individual model performance. 
Works including Mixture-of-Agents \citep{wang2024mixtureofagentsenhanceslargelanguage}, LLM Blender \citep{llm-blender-2023}, and orchestration frameworks like DSPy \citep{khattab2023dspy} demonstrate these approaches' effectiveness. 
Even with single models, techniques like chain-of-thought \citep{wei2023chainofthoughtpromptingelicitsreasoning} and branch-solve-merge \citep{saha2024branchsolvemergeimproveslargelanguage} enhance reasoning capabilities.
Our work extends this literature by focusing specifically on trained-verifier-free inference-time scaling methods that don't require additional reward models. 
Our findings align with and extend \citet{wang2024mixtureofagentsenhanceslargelanguage}, showing that majority voting—especially when weighted by reasoning length—achieves the best balance between computational cost and performance. By constructing comprehensive efficiency–performance trade-off curves, we reveal three key insights: (1) non-reasoning models consistently underperform reasoning-specialized models, even under large inference-time budgets; (2) more sophisticated methods such as sequential revisions offer minimal gains over simple majority voting for reasoning models; and (3) response features, including length and linguistic markers, provide reliable signals for identifying high-quality outputs.

\subsection{Response-Level Signals of Correctness}

Beyond allocating additional inference-time compute, prior work has
investigated whether a model's output contains signals of its own
correctness. Early calibration studies examined whether token
probabilities align with answer accuracy
\citep{jiang-etal-2021-know,tian-etal-2023-just}. Related work on
linguistic calibration studies whether expressions of confidence and
uncertainty accurately reflect the factual reliability of generated
responses \citep{mielke-etal-2022-reducing}. These studies show that
model confidence can be informative, but is not consistently calibrated
across models and tasks. Reasoning traces provide additional response-level signals. Recent work
has examined the relationship between reasoning length and correctness,
finding that models may overthink some problems while underthinking
others \citep{su-etal-2025-underthinking}. Other studies find that
lexical indicators of uncertainty, including hedging expressions, can
provide stronger and more transferable correctness signals than trace
length alone \citep{vanhoyweghen-etal-2025-lexical}. Our analysis complements this literature in two ways. We first compare
the relationship between response length and correctness across both
general-purpose and reasoning-optimized models. We then examine whether
normalized discourse, hedging, and thinking markers distinguish correct
from incorrect reasoning traces. These features require no additional
model generations and can therefore provide lightweight signals for
response selection and diagnosis.

\section{Methodology}

\subsection{Models}
The study evaluates a diverse set of models to cover a wide range of model sizes and architectures, crucial for understanding the effectiveness of ITC methods across different capabilities. The models are categorized into non-reasoning and reasoning models, reflecting their primary strengths and training focus.

\paragraph{Non-reasoning models} Non-reasoning models are general-purpose LLMs optimized for tasks like text generation and dialogue, but they may lack specialized training for complex reasoning. The selected models include: GPT-4o-mini, Qwen2.5-7B-Instruct, Qwen2.5-72B-Instruct \citep{qwen}, Llama-3.3-70B-Instruct, and Llama-3.1-8B-Instruct \citep{dubey2024llama3herdmodels}. This selection covers a wide range of sizes (from 8B to 72B parameters) and includes both open-source and closed-source models, ensuring a comprehensive evaluation. The inclusion of GPT-4o-mini as a closed-source model contrasts with the open-source Qwen2.5 and Llama series, highlighting the study's intent to assess performance across different accessibility models.


\paragraph{Reasoning models} Reasoning models are specifically trained or designed to handle complex reasoning tasks, such as mathematical problem-solving and code generation, often through methods like reinforcement learning (RL). The selected models include: DeepSeek-R1-Distill-Llama-70B \citep{deepseek-ai2025deepseek0r10}, DeepSeek-R1-Distill-Llama-8B, DeepSeek-R1-Distill-Qwen-32B, DeepSeek-R1-Distill-Qwen-14B, DeepSeek-R1-Distill-Qwen-7B, and QwQ-32B-Preview \citep{qwq-32b-preview}. For some experiments we also evaluate DeepSeek-R1 \citep{deepseek-ai2025deepseek0r10}, but not all due to the high-cost nature of inference-scaling methods.

\begin{figure*}[t]
    \centering
    \includegraphics[width=\linewidth,trim={5cm 11cm 14cm 2cm},clip]{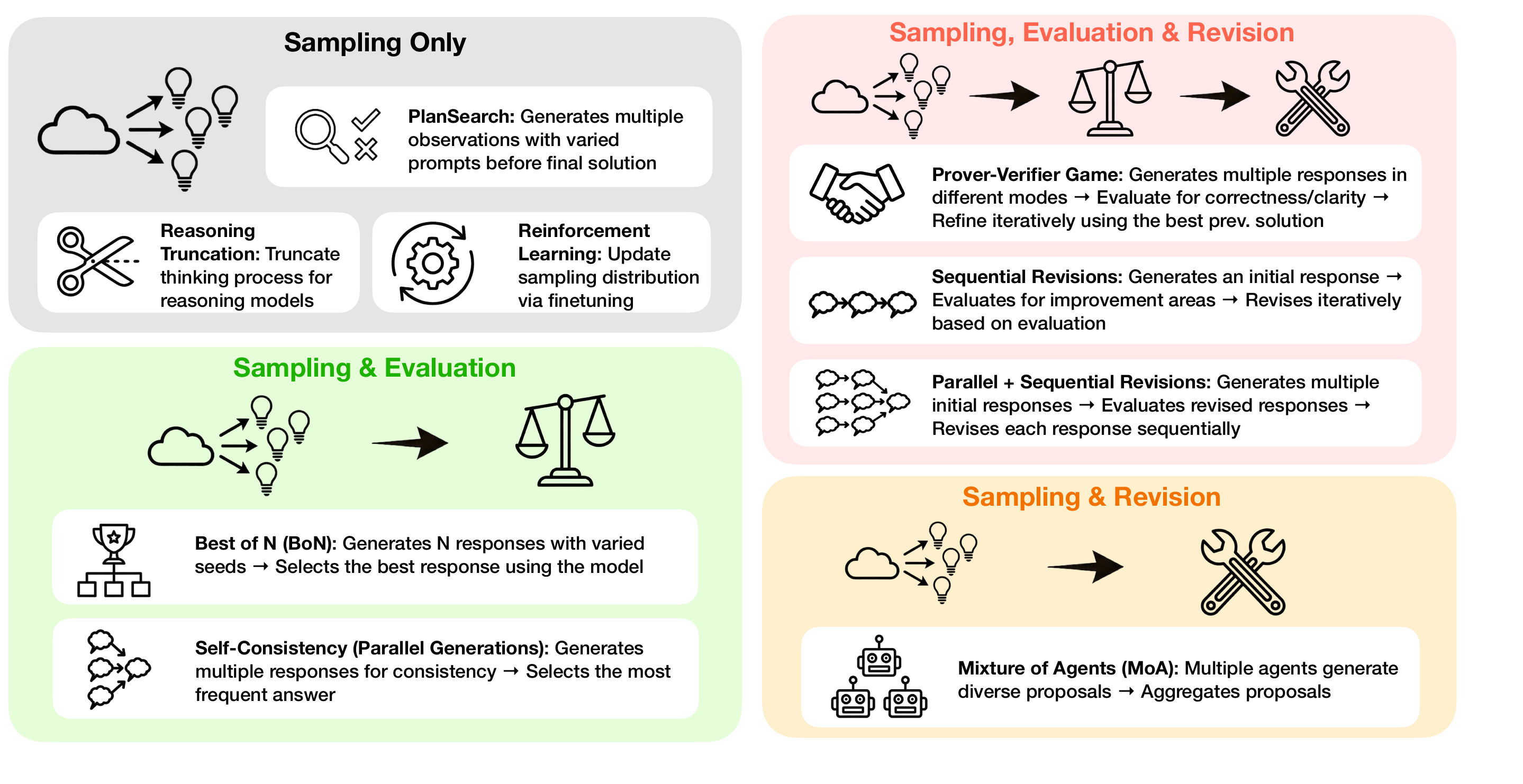}
    \caption{Inference-time scaling methods: sampling, evaluation, and revision approaches. Note that we view reinforcement learning and reasoning truncation as an approach to change the sampling distribution.}
    \label{fig:method_overview}
\end{figure*}

\subsection{Tasks}

\paragraph{MATH 500} MATH 500 is a subset of challenging competition mathematics problems from the MATH dataset  \citep{hendrycksmath2021}. The dataset contains complex high school math problems that are often solved with step-by-step reasoning.

\paragraph{AIME} AIME consists of problems from the American International Mathematics Examination (AIME), a prestigious mathematics competition for high school students. AIME is known for its difficult and thought-provoking problems and we select the 2024 subset as our evaluation benchmark which contains 30 problems.

\paragraph{GPQA} Graduate-Level Google-Proof Q\&A Benchmark (GPQA) \citep{rein2023gpqa}, is a dataset of 448 multiple-choice questions in biology, physics, and chemistry, crafted by domain experts. It is designed to be extremely challenging, with experts who have or are pursuing PhDs in the corresponding domains reaching only 65\% accuracy. We use GPQA Dimaond which is a high quality subset.

\paragraph{LiveCodeBench} Livecodebench \citep{jain2024livecodebench} provides a holistic and contamination-free assessment of large language models for code-related tasks. The code generation (codegen) subtask, specifically, is selected in this work to test the models' ability to generate correct code for these problems. We use the code problems from 2024-11-01 to 2025-02-01.

\subsection{Inference-Time Scaling Methods}

Inference-time scaling methods in language models leverage additional computational resources during the inference phase to enhance performance by adaptively modifying the model’s output distribution for a given prompt at test time. This process involves altering how responses are generated and processed to achieve more accurate or complex outputs compared to direct sampling from the model. The study categorizes these methods into three key steps: sampling, evaluation, and revision, each defined as follows:

\paragraph{Sampling} is the process of generating one or more responses from the language model for a given prompt, potentially with modifications to the input to influence the distribution from which samples are drawn. In inference-time scaling, sampling goes beyond simply drawing responses from the model’s default distribution. Modifications at the input level—such as adding specific tokens or phrases (e.g., in chain-of-thought prompting)—shift the distribution to favor responses that align with desired reasoning outcomes.

\begin{figure*}[t]
    \centering
    \includegraphics[width=\linewidth]{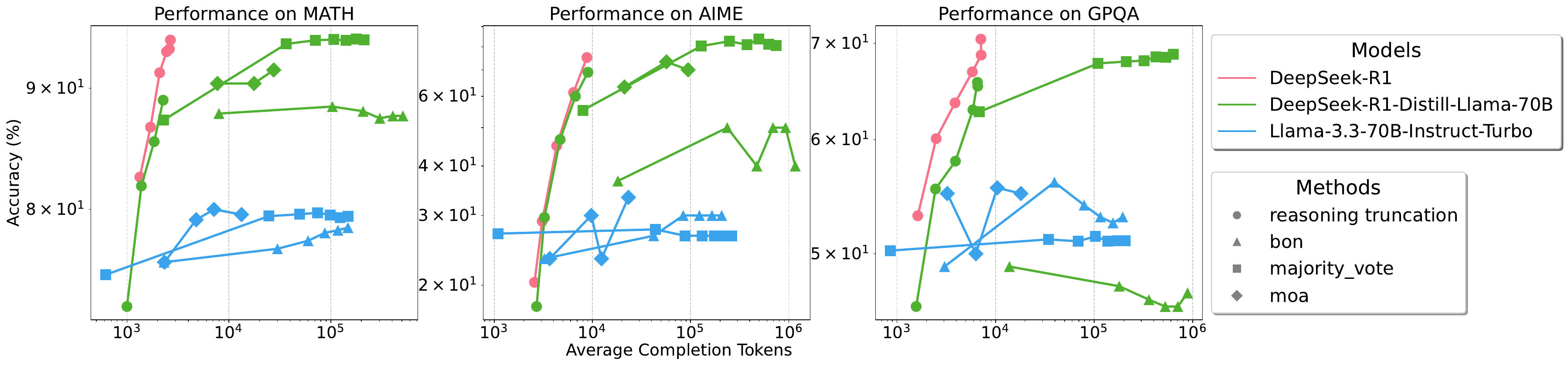}
    \caption{The overview of inference-time-compute methods for reasoning and non-reasoning models. Even though inference-time scaling method improves Llama-3.3-70B, it still struggles to beat the R1-distilled version of Llama 70B. However, with very limited compute, non-reasoning model with inference method can be at the trade-off curve.}
    \label{fig:pareto_front}
\end{figure*}

\paragraph{Evaluation} is the process of assessing the quality or correctness of the generated responses, which informs how we modify the output distribution by selecting, weighting, or ranking the responses. Once responses are sampled, evaluation determines their value, often using metrics like accuracy, coherence, or task-specific scores given by human, reward models or LLMs. In the context of inference-time scaling, this step shapes the output distribution by concentrating it on higher-quality responses. In this work all verifiers used are LLMs, meaning the same model is used for both sampling and evaluation, which can introduce biases but enhances generalizability without requiring external rewards.

\paragraph{Revision} is the process of modifying or improving the generated responses based on the evaluation, potentially involving iterative refinement or generation of new responses, thereby transforming the output distribution to concentrate on higher-quality outputs.

We selected several representative methods that enhance language model performance by adaptively modifying the model's output distribution at test time, using additional compute during inference (\Cref{fig:method_overview}). Each method leverages sampling (generating responses), evaluation (assessing response quality), and revision (refining responses) in distinct ways to achieve this goal. Specifically, we picked PlanSearch \citep{wang2024planning}, Prover-Verifier Game (PVGame) \citep{kirchner2024prover0verifier}, Mixture of Agents (MoA) \citep{wang2025mixtureofagents}, Best of N (BoN), Self-consistency \citep{wang2022self0consistency}, Sequential Revisions, and Parallel + Sequential Revisions \citep{snell2024scaling}. More details can be found in the Appendix.

\section{The Limits of Inference-Time Scaling}

In this section, we will give an overview of various inference-time scaling methods and present the current trade-off curve for those methods. We aim to provide a comprehensive view and provide guidelines for what model or method to use with a given inference budget.

\subsection{The ``Reasoning Floor'': Between Test-Time Scaling and Train-Time Scaling}

\begin{figure}[t]
    \centering
    \includegraphics[width=0.75\linewidth]{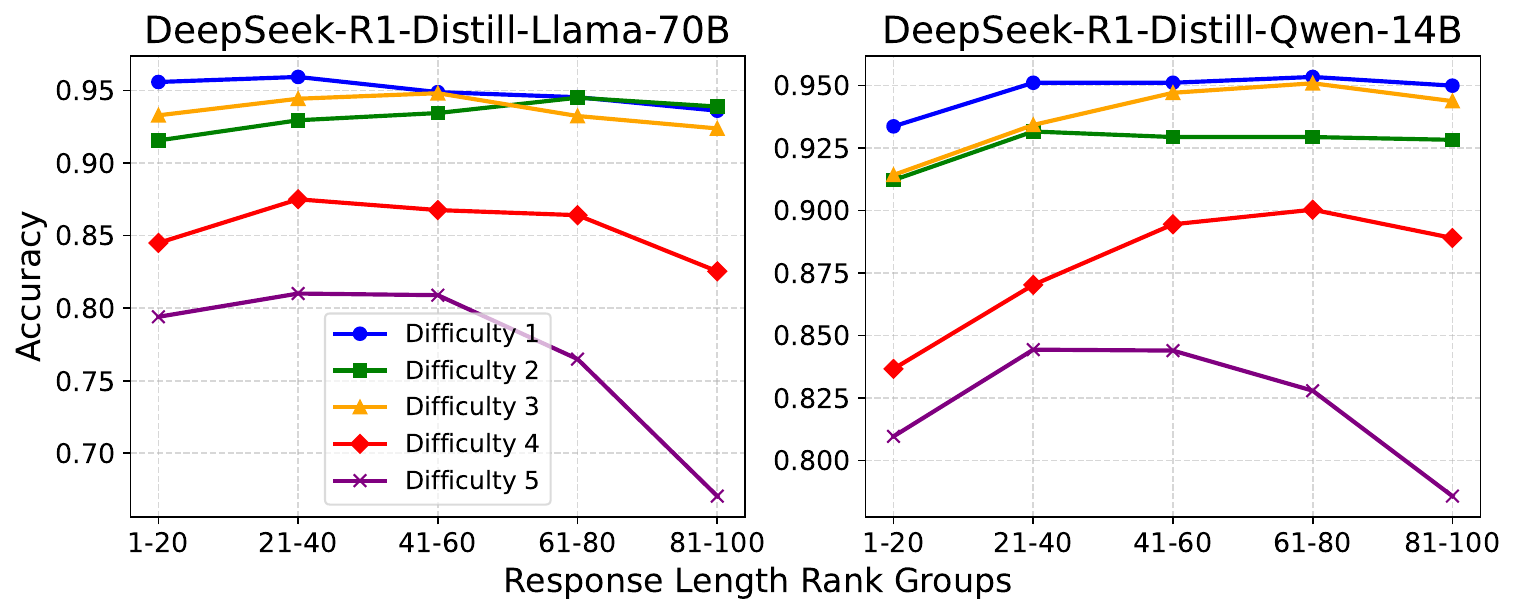}
    \caption{Accuracy of responses in different length groups for various difficulty. For each question, we generate 100 samples and then we bin those samples into five bins. Then average accuracy is computed for each bin across the dataset.}
    \label{fig:rank_vs_accuracy_difficulty}
\end{figure}

\begin{figure*}[t]
    \centering
    \includegraphics[width=1\linewidth]{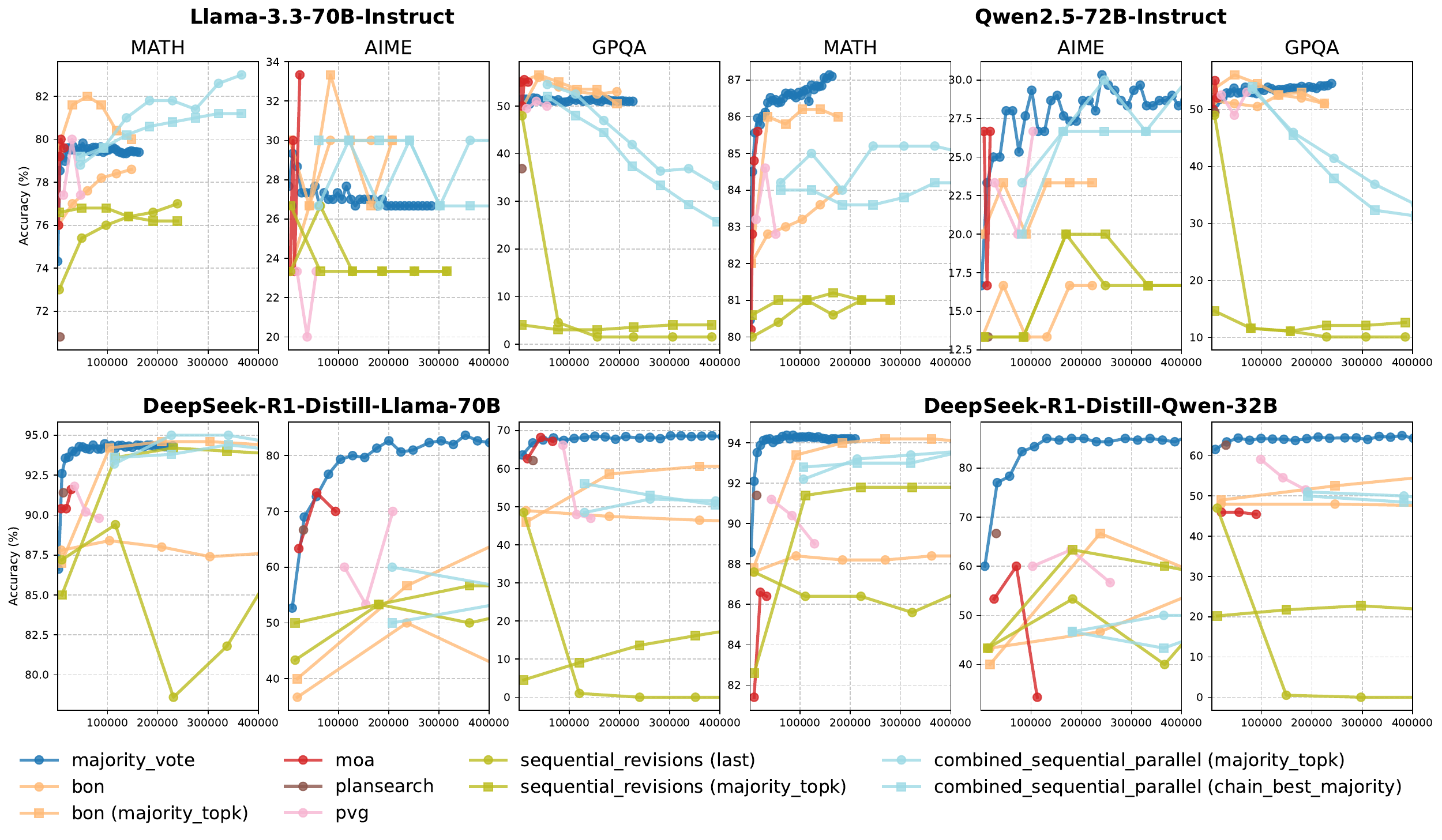}
    \caption{Performance of various inference-time scaling methods for four models across MATH, AIME and GPQA. For some methods we have multiple eval metrics. For \emph{bon} (best-of-N approach), we pick the highest scored response. \emph{majority topk} means we use top k scored resposnes to do majority voting (we always set k as half of total samples). \emph{last} means we  pick the last revisioned sample. \emph{chain best majority} indicates that we use the best scored sample from each chain and then take majority. Due to high cost of inference, methods like sequential revisions and combined sequential parallel are only sampled once, which may seem volatile when plotted. The results for other models can be found in the Appendix.}
    \label{fig:inf_methods_4models_3tasks}
\end{figure*}

While prior work has suggested that increased inference-time compute can substitute for model parameters \citep{snell2024scaling}, \Cref{fig:pareto_front} identifies a distinct ``reasoning floor'': a performance plateau that general-purpose models cannot escape, regardless of the search budget. Specifically, we find that reasoning-optimized models (e.g., R1-Distill-Llama-70B) demonstrate superior token efficiency, achieving peak accuracy with orders of magnitude less compute than their non-reasoning counterparts like Llama-3.3-70B-Instruct (Similar trend is observed in \Cref{fig:inf_methods_4models_3tasks}). Even when general models are provided with an expansive inference budget ($N=256$), they fail to bridge the performance gap to reasoning-optimized weights. This suggests that internalizing reasoning protocols via reinforcement learning is a more effective test-time scaling; without these internalized weights, external search methods yield diminishing returns.

To further isolate the value of these internalized protocols, we introduced a "reasoning truncation" baseline. By force-terminating the reasoning process (within the $\langle$think$\rangle$ blocks) and prompting for immediate sequence completion, we observed a catastrophic degradation in response quality. Interestingly, as the reasoning budget is aggressively truncated, the performance of reasoning models eventually regresses to the "floor" occupied by non-reasoning models. This suggests that the latent "thought" tokens are not merely a byproduct of scale, but the primary vehicle through which compute is converted into logic.

\subsection{Internalized Reasoning vs. Externalized Search Saturation}

Training-free, verifier-free inference-time scaling methods offer minimal improvements for reasoning models. As shown in \Cref{fig:inf_methods_4models_3tasks}, these sophisticated methods fail to significantly enhance the performance of reasoning models. Almost all the methods are underperforming majority voting for both DeepSeek-R1-Distill-Llama-70B and DeepSeek-R1-Distill-Qwen-32B. We hypothesized excessive external revision often introduces \textit{reasoning drift}, where the model's high-probability internal path is distracted by sub-optimal external feedback. 

Conversely, non-reasoning models (e.g., Llama-3.3-70B-Instruct) show greater receptivity to externalized search, occasionally surpassing majority voting. This suggests that for models without specialized weights, external loops can partially simulate the reasoning protocols established in specialized architectures \citep{snell2024scaling}. However, as established by the Reasoning Floor, these gains are ultimately capped by the model's underlying weights. We conclude that for reasoning-optimized models, the most compute-efficient strategy is to prioritize sample diversity over search complexity.

\subsection{Majority Voting as the Compute-Optimal Baseline}

Across both reasoning and non-reasoning models, majority voting consistently demonstrates superior performance. The method's simplicity belies its effectiveness, as illustrated in \Cref{fig:inf_methods_4models_3tasks}, where it outperforms more complex inference-time scaling approaches such as mixture-of-agents, best-of-N, sequential revisions and combined parallel sequential.
The success of majority voting can be attributed to its fundamental approach of leveraging multiple model outputs. By aggregating responses, the method effectively mitigates individual model biases and captures a more robust representation of the underlying reasoning. 

\begin{figure*}[t]
    \centering
    \includegraphics[width=0.85\linewidth]{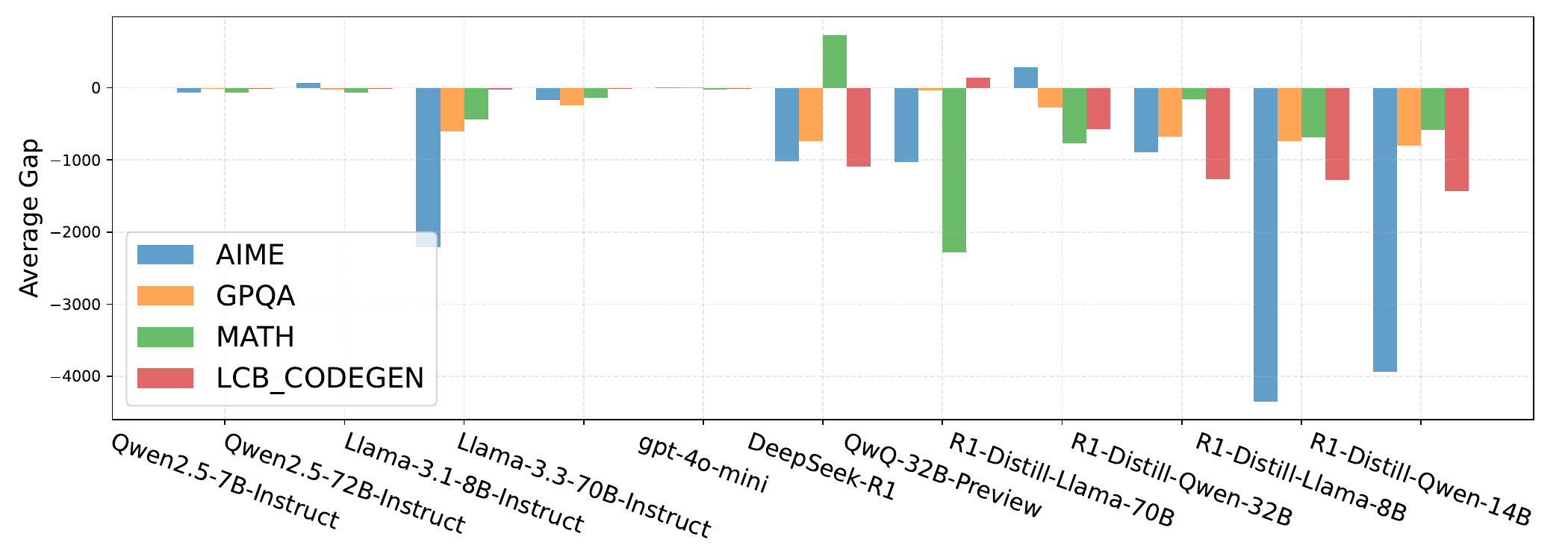}
    \caption{The average response length gap for each model tasks across four tasks. The average response length gap is computed by: 1) calculating mean length difference between correct and incorrect responses within each question and 2) averaging these differences across the entire dataset. LCB\_CODEGEN represents the code generation subtask in the LiveCodeBench benchmark.}
    \label{fig:length_gap_histogram}
\end{figure*}

\subsection{Scaling Behavior vs. Problem Difficulty}

\Cref{fig:length_gap_histogram} reveals a critical insight into model performance across varying task complexities. As problem difficulty increases, the gap between correct and incorrect responses becomes more pronounced, particularly for reasoning models. The AIME dataset, known for its challenging nature, exemplifies this trend, with all reasoning models demonstrating a wider correctness gap. To systematically investigate this phenomenon, we analyze response length differences using the MATH dataset, which offers a natural difficulty gradient ranging from level one to five. We stratify samples for each question into five bins ranked from shortest to longest responses.
We find that for high-difficulty problems (level 5), an inverse relationship between response length and correctness becomes evident. Specifically, \Cref{fig:rank_vs_accuracy_difficulty} demonstrates that as problem complexity increases, reasoning models tend to generate more accurate responses with shorter lengths.

\section{Diagnostic Signals in Reasoning Traces}
\subsection{The Impact of Response Length to Performance}


Our analysis reveals a distinct divergence between model types. While general-purpose models show no stable correlation, reasoning-optimized models demonstrate a significant inverse relationship: shorter, more direct responses are more likely to be correct.

\subsubsection{Non-reasoning Models: No Correlation}

Non-reasoning models exhibit no discernible correlation between response length and accuracy. \Cref{fig:length_gap_histogram} plots the average response length gap for each model tasks across four tasks. The average response length gap is computed via first the mean length difference between correct and incorrect responses within individual
questions (There are 100 samples for resoning models and 256 samples for non-reasoning models). Then we average these differences across the entire dataset. We discard the questions that only include one class of correctness (all correct, or all incorrect). This ensures that the average gap is not biased by questions that are too easy or too difficult. It reveals that the response length gaps across all four datasets are low for all non-reasoning models, with an exception of Llama-3.1-8B-Instruct. In that case, we observe a non-negligible gap for the AIME task.

\subsubsection{Reasoning Models: Inverse Correlation}

Unlike non-reasoning models, reasoning models show a clearer trend of shorter, more precise responses being more accurate. \Cref{fig:length_gap_histogram} provides clear evidence of this inverse relationship between response length and accuracy. This suggests that for reasoning-optimized weights, correctness is fundamentally tied to efficient reasoning paths. In contrast, incorrect outputs often involve redundant logic, circular thinking, or failed self-correction cycles that artificially inflate the response length. This divergence indicates that in specialized models, verbosity serves as a diagnostic signal for reasoning failure.

\subsection{Linguistic Markers and Word Frequency Analysis}

\begin{figure*}
    \centering
    \includegraphics[width=0.9\linewidth]{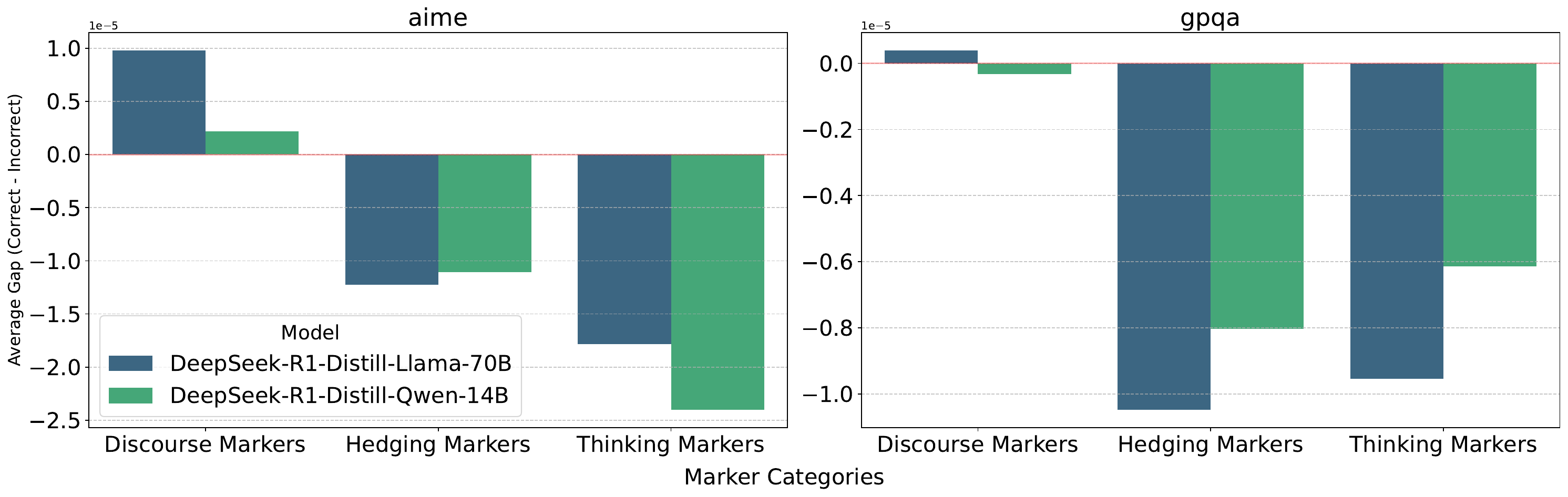}
    \caption{Average gaps between correct and incorrect responses for DeepSeek-R1-Distill-Llama-70B and DeepSeek-R1-Distill-Llama-14B. The average gaps are first computed by using computing the mean difference of thinking token frequency of correct and incorrect responses within one question and then average over the entire dataset. The frequency is weighted by response length. Refer to \Cref{tab:linguistic_markers} for the definition of different marker categories.}
    \label{fig:marker_category_summary_all_models_tasks}
\end{figure*}


Reasoning models tend to use certain linguistic markers, especially thinking tokens such as ``alternatively'' or ``however''. In this section, we investigate the relationships between such linguistic markers and correctness. 

\subsubsection{Linguistic Markers: Higher Frequency in Incorrect Responses}

A compelling finding emerges from our linguistic marker analysis: incorrect responses consistently exhibit a higher density and diversity of linguistic markers. \Cref{fig:marker_category_summary_all_models_tasks} provides empirical evidence that both hedging and thinking markers (marker definition can be found in \Cref{tab:linguistic_markers}) are markedly more prevalent in incorrect responses compared to correct ones.
Our methodological approach involved computing the average marker frequency gaps by first calculating the mean difference in thinking token frequencies between correct and incorrect responses within individual questions, and then averaging these differences across the entire dataset. To account for response variability, we normalized the marker frequencies by response length. Note that this normalization will naturally result in very low frequency. Normally the marker counts range between 5 and 30 per response.
The proliferation of markers in incorrect responses suggests a fundamental characteristic of ineffective reasoning. Rather than representing depth or complexity, the increased marker density appears to be a signal of cognitive imprecision—a tendency toward verbose and less focused responses. This observation provides a promising diagnostic tool for identifying potentially incorrect model outputs, suggesting that the quality of reasoning can be partially assessed through the strategic use and density of linguistic markers. The list of extra markers can be found at \Cref{tab:linguistic_markers} in the Apprendix.

\subsubsection{Linguistic Markers as a Correctness Predictor}

The notable discrepancy in marker counts between correct and incorrect responses motivated a further investigation into the predictive potential of linguistic markers. We designed a rigorous experimental framework to explore this hypothesis.
Our method involved creating a dataset with extensive model response samples, generating 100 response generations per question, and then training a accuracy classifier based on the response features. We extracted marker features as input variables and response correctness as the target label. To ensure the robustness of model evaluation, we implemented a standard 0.6/0.2/0.2 train-validation-test split.
The results show great promises. For the DeepSeek-R1-Distill-Llama-70B model, a simple logistic regression classifier achieved a test F1 score of 0.7469. The performance was even more impressive for the DeepSeek-R1-Distill-Llama-14B model, with an F1 score of 0.8637.
Our finding suggests the potential of linguistic markers as a diagnostic tool for assessing model reasoning capabilities. The approach offers a nuanced method for evaluating model performance beyond traditional length metrics, opening new avenues for model interpretation and quality assessment.

\section{Conclusion}
Our work thoroughly assesses trained-verifier-free inference-time scaling methods for LLMs, emphasizing their efficiency and effectiveness in reasoning tasks. We identify a fundamental ``reasoning floor'': a performance ceiling where general-purpose models, despite leveraging advanced scaling techniques and significant computational resources, fail to match the baseline accuracy of specialized reasoning architectures. For reasoning models, simpler strategies such as majority voting often surpass more intricate methods like best-of-N or sequential revisions in performance. Furthermore, our analysis reveals that correct reasoning traces are characterized by higher efficiency—typically appearing as shorter responses with a lower density of linguistic markers such as hedging and thinking tokens. These intrinsic features serve as zero-compute diagnostic signals for accuracy. Leveraging these linguistic markers to develop self-correcting agents and more efficient aggregation methods represents a promising frontier for advancing stable, complex reasoning.

\section{Limitations}
While this work establishes the concept of ``reasoning floor'' and identifies key linguistic signals for RM-free scaling, a few limitations remain. First, our analysis focuses on trained-verifier-free methods. While this ensures generalizability across domains without specialized reward models, the inclusion of process-reward models (PRMs) or outcome-reward models (ORMs) might shift the Pareto front, potentially allowing non-reasoning models to bridge the efficiency gap more effectively. Second, our linguistic signal of correctness can be evaluated on open-ended agentic tasks, where verbosity might be a requirement for task completion.

\bibliographystyle{plainnat}
\bibliography{custom}

\newpage

\appendix

\begin{table*}
    \centering
    \renewcommand{\arraystretch}{1.4} 
    \setlength{\tabcolsep}{12pt} 
    \begin{tabularx}{\textwidth}{lX}
        \toprule
        \rowcolor{lightgray} \textbf{Category} & \textbf{Examples} \\
        \midrule
        \textbf{Discourse Markers} & \textit{on the other hand}, \textit{nevertheless}, \textit{moreover}, \textit{in addition}, \textit{furthermore}, \textit{therefore}, \textit{consequently}, \textit{as a result} \\
        \rowcolor{lightgray} \textbf{Hedging Markers} & \textit{perhaps}, \textit{maybe}, \textit{possibly}, \textit{it seems}, \textit{might}, \textit{could} \\
        \textbf{Thinking Markers} & \textit{however}, \textit{wait}, \textit{alternatively}, \textit{hmm} \\
        \bottomrule
    \end{tabularx}
    \caption{Categories of Linguistic Markers. Italicized words represent specific markers.}
    \label{tab:linguistic_markers}
\end{table*}
\section{Usage of LLM}
We utilize LLM to assist some of the paper writing, code development and figure generation.

\section{Inference-Time Scaling Methods}
\label{appendix:inf-methods}

In this section, we will go into more detils about each inference-time scaling method and how they are parametrized in our studies.

\paragraph{Majority Vote} Majority vote (self-consistency) generate multiple samples and choose the most frequent answer as final solution. Note that this method doesn't quite work for free-form generation problems such as LiveCodeBench. Hence we don't present results for LiveCodeBench. For reasoning model, we sample 100 samples for each query while for non-reasoning models, we sample 256 samples. For reasoning models, 100 samples were used due to higher computational cost and our observation that performance trends stabilized at this point. Non-reasoning models, being less computationally intensive, used 256 samples for greater robustness.

\paragraph{PlanSearch} This method prompts LLM to generate a number of observations and derived observations before making a final solution. We set number of generations to be three and number of derived observations to be two across all experiements. 

\paragraph{PVGame} Prover-Verifier Games involves two main phases: solution generation and solution verification. Solutions are generated in both "helpful" and "sneaky" modes, evaluated for correctness and clarity, and refined iteratively by leveraging the best-scored solutions from prior rounds to guide subsequent attempts. In our setup, we fix the number of solutions each round to be three, and scale the number of rounds from one to three.

\paragraph{Best of N} Best of N samples N generations and each generation is evaluted via a judge. We use LLM as a judge and the judge template is shown in LLM evaluator prompt section below. We evaluate three times for each question and then take the mean as the final score. 

\paragraph{Sequential Revisions} We follow the implementation of \cite{snell2024scaling}. This method samples solution sequenially and each time it is revising from last solution except the first solution. The revision process involves first prompting feedback and then asking LLM to provide a revision based on those feedbacks. The feedback prompt template and the revision template first prompt is detailed in the LLM revision feedback prompt section and LLM revision prompt below respectively. Each solution is also evaluted using LLM as a judge. This is for choosing the final solution from the samples.

\paragraph{Parallel + Sequential Revisions} We again follow the implementation of \cite{snell2024scaling}. This method samples multiple generations in the first step. For each geneneration, it sequentially revise similar to sequential revision independently. Same prompts are used from sequential revisions. 

\paragraph{Reasoning Truncation} We truncate the reasoning process (encapsulated in ``$\langle$ think$\rangle$" ``$\langle$/think$\rangle$" tokens) to control for the budget. 

\section{Prompts Templates}
\subsection{LLM Evaluator Prompt}
\begin{tcolorbox}[colback=blue!5!white, colframe=blue!25!black, title=LLM evaluator prompt, width=\columnwidth]
Question: \textbf{\{question\}}

Response: \textbf{\{response\}}
\\ \\
Analyze this answer strictly and critically. Identify and point out every flaw and imperfection to deduct the appropriate amount of points. Be very harsh and stringent in your assessment to ensure the grades are authoritative and reliable. Never award full marks. Assign a score between -100 and +100.
\\ \\
Response format: 

[Analysis] ...

[Score] a single integer between -100 and +100.
\label{tab:judge_template}

\end{tcolorbox}
\subsection{LLM Revision Feedback Prompt}
\begin{tcolorbox}[colback=blue!5!white, colframe=blue!25!black, title=LLM revsion feedback prompt, width=\columnwidth]
Since we have a weak Answer, could you provide me with a relection or feedback to correct this answer better? Analyze this Answer Strictly and Critic, point out every flaw for ervery possible imperfect to minus every possible score!
Let's think step by step.
\\ 
\\
Question: \textbf{\{question\}}
\\
\\
Answer to analyze:
\textbf{\{previous response\}}
\label{tab:revision_feedback_template}
\end{tcolorbox}
\subsection{LLM Revision Prompt}
\begin{tcolorbox}[colback=blue!5!white, colframe=blue!25!black, title=LLM revision prompt, width=\columnwidth]
Please refine the your answer according to your Reflection or Feedback. The response should begin with [reasoning process]...[Verification]... and end with end with "[Final Answer] The answer is \\boxed\{answer\}"
\\
Let's think step by step.
\\
\\
Question: \textbf{\{question\}}
\\
\\
Previous solution:
\textbf{\{previous response\}}
\\
\\
Feedback: \textbf{\{feedback\}}
\label{tab:revision_template}
\end{tcolorbox}

\section{More Inference Time Scaling Results}
In this section, we present results for more models: Meta-Llama-3.1-8B-Instruct, Qwen2.5-7B-Instruct, gpt-4o-mini, DeepSeek-R1-Distill-Qwen-14B (Figure 6).
\label{appendix:more_inf_results}
\begin{figure*}[t]
    \centering
    \includegraphics[width=1\linewidth]{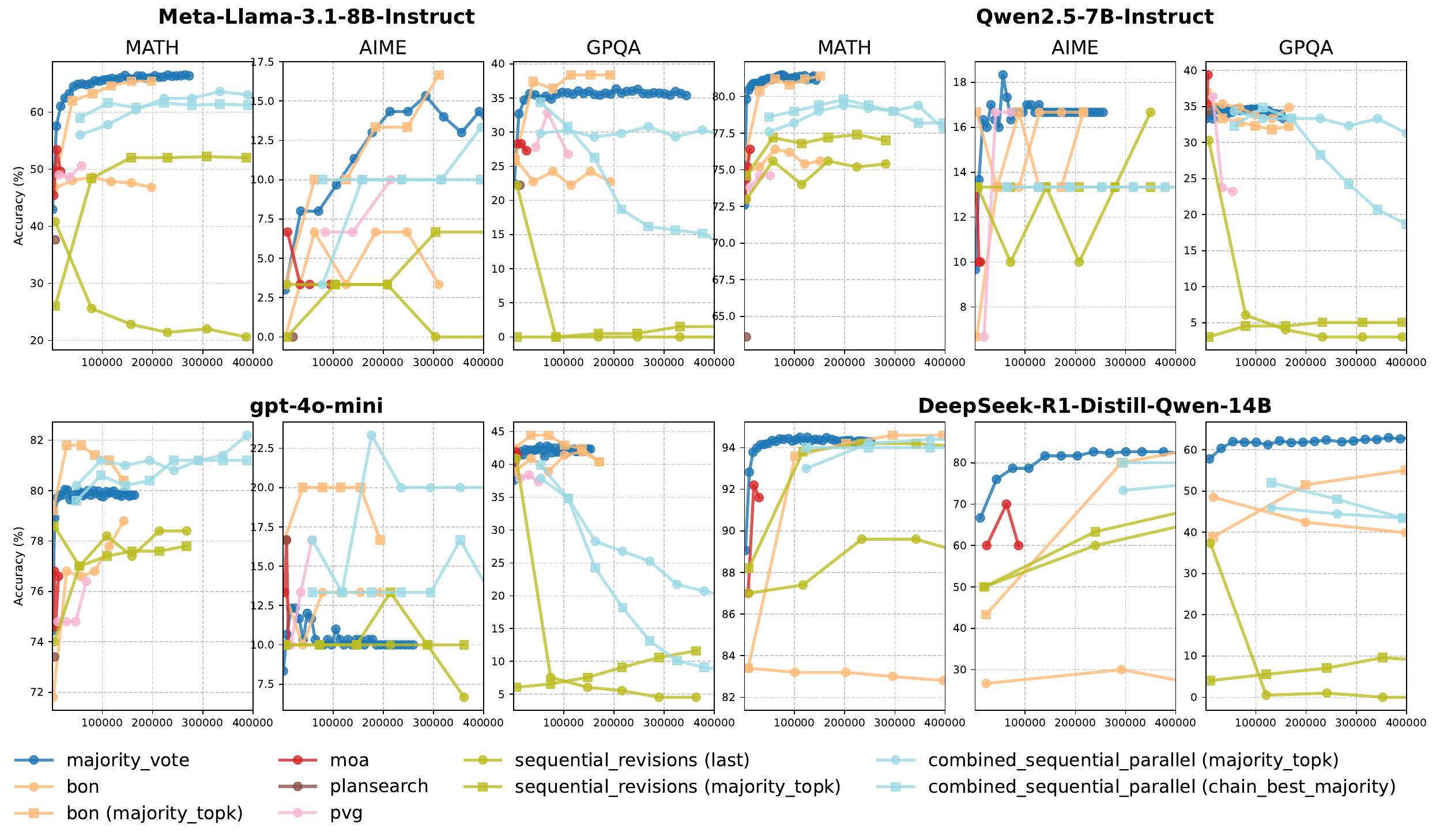}
    \caption{Performance of various inference-time scaling methods for four models across MATH, AIME and GPQA. }
    \label{fig:inf_methods_4models_3tasks2}
\end{figure*}

\section{Inference Budget}
All runs are conducted on 8xA100 or through Together API. 
\end{document}